\documentclass{article}

\PassOptionsToPackage{table}{xcolor}
\usepackage{xcolor}
\usepackage[preprint]{corl_2026}
\usepackage{enumitem}
\usepackage[utf8]{inputenc}
\usepackage[T1]{fontenc}
\usepackage{url}
\usepackage{booktabs}
\usepackage{amsfonts}
\usepackage{nicefrac}
\usepackage{microtype}
\usepackage{amsmath}
\usepackage{amssymb}
\usepackage{graphicx}
\usepackage{longtable}
\usepackage{multirow}
\usepackage{array}
\usepackage{bm}
\usepackage{tikz}
\usetikzlibrary{positioning, arrows.meta, fit, backgrounds, calc, shapes.geometric}
\usepackage{makecell}
\usepackage{wrapfig}
\usepackage{capt-of}
\usepackage{subcaption}
\usepackage{float}

\graphicspath{{./}{figs/}{../figs/}}

\hypersetup{
  pdftitle={KAM-WM: Kinematic Affordance Maps from Latent World Models for Robot Manipulation},
  pdfauthor={Xinyu Shao, Keru Zhou, Guowei Huang, Yajun Gao, Tongtong Cao, Xiu Li},
  pdfsubject={Robot manipulation, latent world models, affordance priors},
  pdfkeywords={Robot Manipulation, Latent World Models, Imitation Learning}
}

\title{KAM-WM: Kinematic Affordance Maps from Latent World Models for Robot Manipulation}

\author{
  \textbf{Xinyu Shao$^{1,2}$, Keru Zhou$^{1}$, Guowei Huang$^{2}$, Yajun Gao$^{2}$, Tongtong Cao$^{2}$, Xiu Li$^{1}$}\\[1mm]
  $^{1}$Tsinghua Shenzhen International Graduate School\\
  $^{2}$Huawei Technologies Ltd.\\
  \texttt{shaoxy23@mails.tsinghua.edu.cn}
}

\begin{document}
\maketitle

\begin{abstract}
Learning manipulation from few demonstrations requires visual priors that
capture not only \emph{where} to interact, but also \emph{how} the interaction
should begin; static priors such as segmentation masks encode only the former.
We present KAM-WM, a framework that extracts a coarse directional interaction
cue from a frozen latent video world model without rollout or world-model
fine-tuning. KAM-WM queries a Flow Matching image-to-video backbone once and
interprets its single-step latent velocity as a Kinematic Affordance Map (KAM),
which provides task-conditioned interaction regions and coarse motion
structure. A lightweight Perceiver compresses KAM into tokens that condition a
diffusion policy together with RGB observations and proprioception. Across
LIBERO and RoboTwin~2.0, KAM-WM reaches 90.6\% average success on LIBERO and
achieves 65.7\% and 22.4\% success rates in the Easy and Hard settings on
RoboTwin~2.0, respectively. Controlled comparisons against a zero-order mask
prior suggest that part of the gains comes from directional information beyond
spatial localization alone. These results indicate that, in the evaluated
settings, a frozen video model can provide a useful first-order visual prior for
control without the test-time cost of future rollout.
\end{abstract}


%===============================================================================
\section{Introduction}
\label{sec:intro}

\begin{figure*}[t]
    \centering
    \includegraphics[width=0.97\textwidth]{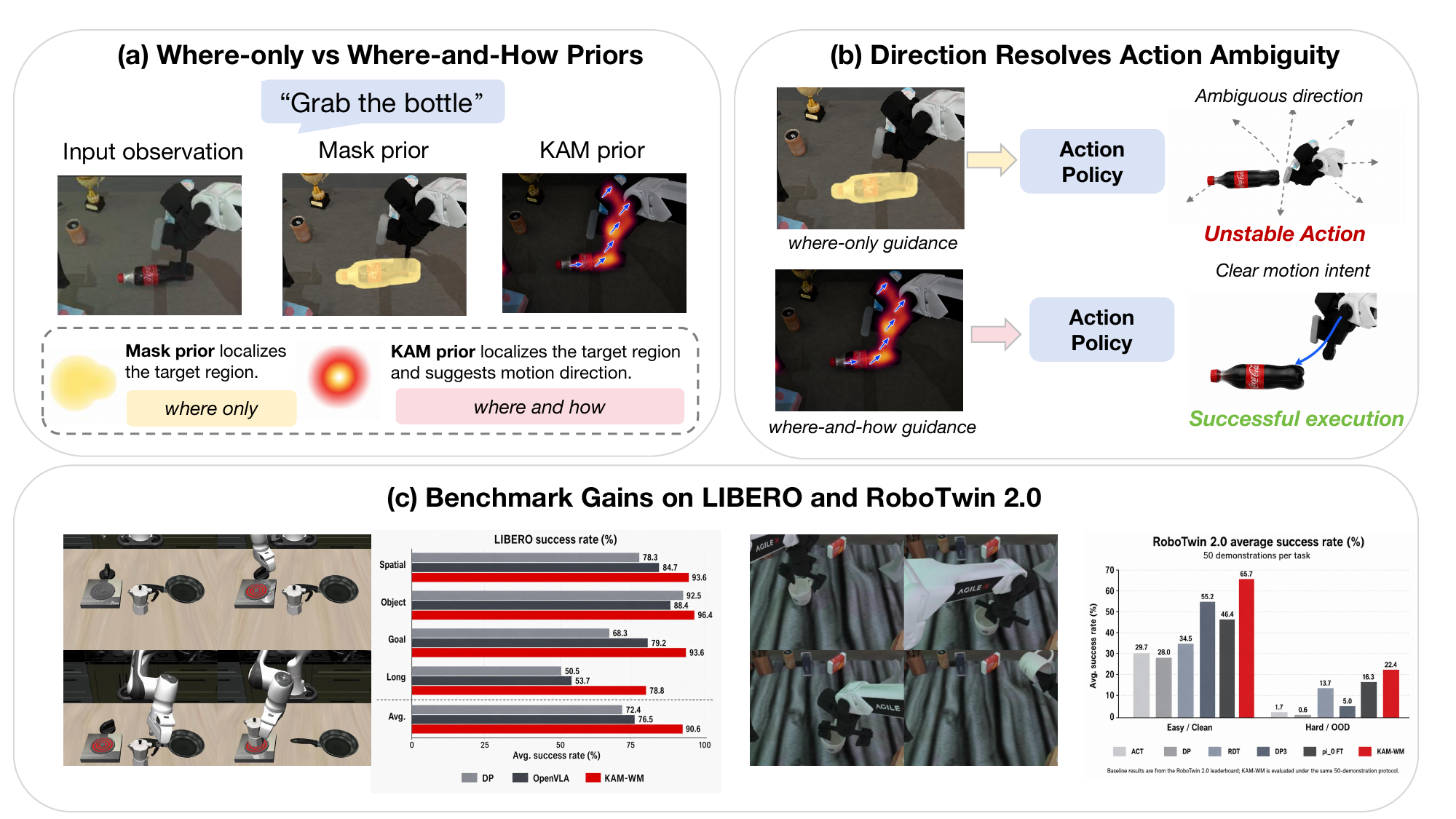}
    \vspace{1mm}
    \caption{\textbf{KAM-WM provides where-and-how cues for low-data
    manipulation.} (a) KAM highlights interaction-relevant regions and motion cues
    beyond object masks. (b) These tokens condition the diffusion policy together
    with RGB and proprioception. (c) KAM-WM improves performance on LIBERO and
    RoboTwin~2.0.}
    \label{fig:teaser}
    \vspace{-6mm}
\end{figure*}

Learning manipulation from few demonstrations requires visual priors that
indicate not only \emph{where} to attend, but also \emph{how} the interaction
should begin. Static priors such as segmentation masks help localization, yet
remain zero-order cues: they mark relevant objects or regions without encoding
approach direction. This is limiting for tasks such as hanging a mug on a hook,
where location alone does not determine a successful motion. As illustrated in
Figure~\ref{fig:teaser}(a), a bottle mask may cover the whole object, whereas a
more useful prior would emphasize the lower graspable region together with the
gripper's approach motion.

Pretrained video models are an appealing source of this directional information,
since future-frame prediction requires encoding how objects move and interact.
Most robotics methods that use this knowledge, however, roll out future frames
as visual subgoals or co-train the backbone with actions. Both add cost:
iterative denoising increases test-time latency, and fine-tuning a large video
model is expensive in the low-data regime we target. This raises a natural
question: can motion knowledge in a frozen video model be read off directly,
without generating future frames?

Our starting point is a simple empirical observation: when a frozen
text-conditioned Wan~2.2 image-to-video model is queried at the first denoising
step, its high-response regions often align with task-relevant contact structure
and the robot embodiment. In a bottle manipulation scene, for example, the
response highlights the lower graspable region and approaching arm, consistent with
both sharper localization and a coarse interaction cue.

% --- Teaser ---

This observation is consistent with the Flow Matching training
objective~\cite{lipman2022flow,liu2022flow}. For a conditional video model, the
single-step latent velocity at the high-noise endpoint ($t{=}1.0$) estimates a
model-dependent displacement toward plausible future video latents conditioned
on the current observation and language instruction; we use this field as a
\textbf{Kinematic Affordance Map (KAM)}, whose magnitude highlights
task-conditioned response regions and whose normalized latent response provides
structure beyond a binary or soft mask. We do not treat this latent direction as
a metrically calibrated 3D motion field, but as a demonstration-grounded visual
prior that extends zero-order masks while avoiding multi-step rollout and
world-model fine-tuning.

KAM-WM turns this intuition into a control interface. We compress the dense
latent velocity field into 8 tokens with a lightweight Perceiver, and use them
to condition a 1D U-Net Diffusion Policy together with
multi-view RGB observations and proprioception. We do not interpret the latent
field as a direct 3D action plan; KAM is a visual prior rather than an
explicit geometric planner. The policy learns from demonstrations how to align
this latent field with the robot action space.

We evaluate KAM-WM on LIBERO~\cite{liu2023libero} and
RoboTwin~2.0~\cite{chen2025robotwin}. On LIBERO, KAM-WM achieves 90.6\%
average success. On RoboTwin~2.0, under a 50-demonstration protocol, it reaches
65.7\% Easy success across 50 tasks, exceeding the leaderboard ACT, $\pi_0$ FT,
and DP3 Easy averages by 36.0, 19.3, and 10.5 points, respectively. Because
these baselines are taken from the leaderboard, we treat the comparison as
contextual rather than a controlled implementation comparison. Ablations with zero-order
masks and KAM variants suggest that the gains are strongest when interaction
direction matters and reflect information beyond localization alone.

The contributions of this work are:
\begin{itemize}[itemsep=0pt, topsep=1pt, parsep=0pt, partopsep=0pt]
    \item We propose extracting a single-step latent velocity field from a
    frozen Flow Matching image-to-video model as a first-order visual prior
    for low-data manipulation.
    \item We introduce a lightweight Perceiver-token interface that conditions
    a diffusion policy on this dense latent field without video rollout or
    world-model fine-tuning.
    \item We evaluate KAM-WM on LIBERO and RoboTwin~2.0, with ablations
    indicating that the gains are largest on tasks where directional
    interaction cues matter.
\end{itemize}

%===============================================================================
\section{Related Work}
\label{sec:related}

\paragraph{Static visual and affordance representations.}
Visual representation learning has provided strong spatial and semantic features
for robot learning through language supervision, masked reconstruction,
segmentation, and robotics-oriented pre-training on human or embodied
video~\cite{radford2021clip,he2022mae,oquab2023dinov2,kirillov2023sam,nair2022r3m,radosavovic2022real,majumdar2023vc1,bardes2024vjepa}.
Affordance-based methods go further and predict where interaction can occur,
using actionability scores, contact regions, or structured pick-and-place
maps~\cite{mo2021where2act,wu2021vatmart,bahl2023vrb,zeng2021transporternetworks,shridhar2022cliport,shao2025robomap}.
These representations are effective at localizing task-relevant objects, but they
remain largely zero-order: they indicate \emph{what} or \emph{where} to attend
to, while the approach direction and post-contact motion are left for the policy
to infer. KAM-WM is complementary to this line: it reads a first-order velocity
field from a frozen video model, adding a directional cue without requiring
affordance labels.

\paragraph{Video world models and action-conditioned foundation models.}
In robotics, generative video and world models have been used for visual
planning, subgoal-like guidance, and inverse
dynamics~\cite{du2023unipi,hu2024video,black2024susie}. More broadly,
generative models have become a framework for decision-making problems with
structured trajectories~\cite{shao2025generativedecision}.
In parallel, large Vision-Language-Action models learn action-conditioned
representations from large-scale robot
datasets~\cite{brohan2023rt2,team2024octo,black2024pi0,kim2024openvla}. These
models encode rich temporal and semantic priors, but adapting or using them for
low-data manipulation often requires action fine-tuning, action co-training, or
multi-step video rollout, which are costly and add test-time latency. KAM-WM
follows the opposite design point: the video backbone stays frozen and is queried
once at the noise endpoint, so the temporal prior is reused without future-frame
generation or any backbone update.

\paragraph{Motion and flow priors for policy learning.}
A related line of work conditions robot policies on predicted motion, such as
object flow, point trajectories, or any-point tracking, to signal how the scene
may change under an
action~\cite{wen2024atm,bharadhwaj2024track2act,xu2024im2flow2act,yuan2024generalflow}.
These methods show that motion prior can be valuable when static localization is
insufficient, but they typically depend on a dedicated flow or tracking module,
curated trajectory supervision, or an explicit procedure for converting 2D
motion into robot actions. KAM-WM differs in where the motion signal comes from:
rather than training a flow predictor, it extracts a latent velocity field from a
general-purpose video model in a single query and uses it as a coarse visual
prior. We therefore do not provide head-to-head comparisons with
ATM~\cite{wen2024atm}, Track2Act~\cite{bharadhwaj2024track2act},
Im2Flow2Act~\cite{xu2024im2flow2act}, or
GeneralFlow~\cite{yuan2024generalflow}, whose flow or tracking modules are
trained on different data sources and supervision.

%===============================================================================
\section{The KAM-WM Framework}
\label{sec:method}

\begin{figure*}[t]
    \centering
	    \includegraphics[width=\textwidth]{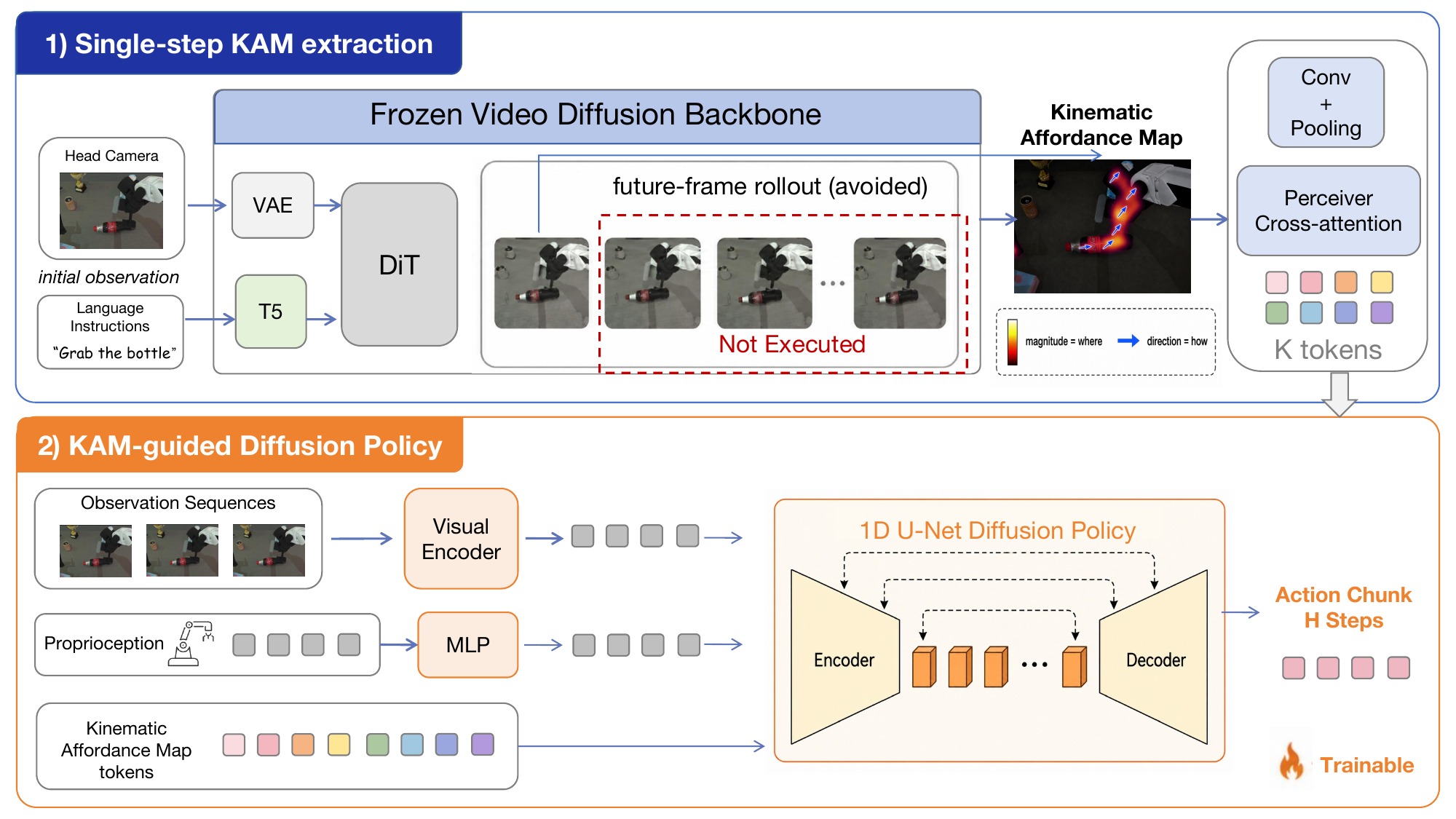}
	    \caption{\textbf{KAM-WM framework.}
      Given the initial observation and language instruction, KAM-WM queries a frozen
      Flow Matching image-to-video backbone once at $t{=}1.0$, with no future-frame
      generation, and reads the resulting single-step latent velocity field as a
      Kinematic Affordance Map. A Perceiver compresses this dense field into $K$
      dynamic tokens, which condition a 1D U-Net Diffusion Policy together with current
      RGB observations and proprioception to predict $H$-step action chunks.}
    \label{fig:framework}
\end{figure*}

KAM-WM has two stages. First, it queries a frozen image-to-video model once at
the start of an episode to obtain a task-conditioned Kinematic Affordance Map
(KAM). Second, it compresses this dense field into a few tokens that condition a
diffusion policy throughout closed-loop execution. The key idea is to use the
model not for video rollout, but for a single-step latent velocity field whose
magnitude highlights task-relevant regions and whose normalized response adds
directional structure for the policy. Figure~\ref{fig:framework} illustrates the
architecture.

%===============================================================================
\subsection{Single-Step Latent Velocity Extraction}
\label{subsec:prelim}

We read a task-conditioned motion cue from a frozen image-to-video model in a
single forward pass, rather than by generating future frames. We use the
official Wan2.2-TI2V-5B release~\cite{wan2025wan} as the backbone, building on
Rectified Flow~\cite{liu2022flow,lipman2022flow}, and query it once at the
maximum-noise endpoint. Let $\mathbf{x}_1 \sim \mathcal{N}(0,\mathbf{I})$ be a
pure-noise latent and $\mathbf{x}_0 \sim q(\mathbf{x}\mid c)$ a clean future
video latent conditioned on $c=[o_1,l]$, with initial observation $o_1$ and
language instruction $l$. We use a denoising-time convention where $t{=}1.0$ is
the noise endpoint and $t{=}0.0$ the data endpoint; Wan~2.2 is trained with
target $\mathbf{v}_{\text{target}} = \mathbf{x}_1 - \mathbf{x}_0$. Under straight
interpolation $\mathbf{x}_t = t\,\mathbf{x}_1 + (1-t)\,\mathbf{x}_0$, a model
trained with squared error estimates
\begin{equation}
    \mathbf{v}_\theta(\mathbf{x}_t,t,c)
    \approx
    \mathbb{E}\!\left[
        \mathbf{x}_1-\mathbf{x}_0
        \mid
        \mathbf{x}_t,c
    \right].
    \label{eq:expected_displacement}
\end{equation}
At $t{=}1.0$, the input is $\mathbf{x}_t=\mathbf{x}_1$, giving
\begin{equation}
    \mathbf{v}_\theta(\mathbf{x}_1,1.0,c)
    \approx
    \mathbf{x}_1
    -
    \mathbb{E}\!\left[
        \mathbf{x}_0
        \mid c
    \right] .
    \label{eq:t1_field}
\end{equation}
With $\mathbf{x}_1$ fixed, variation across $(o_1,l)$ is dominated by the
conditional term $\mathbb{E}[\mathbf{x}_0\mid c]$, so Eq.~\ref{eq:t1_field} is a
conditional latent estimate up to a constant offset. We use it as a visual prior
rather than a calibrated motion field: a single noise-endpoint query depends on
the observation and instruction, with no multi-step denoising.

%===============================================================================
\subsection{Kinematic Affordance Maps}
\label{subsec:kam}

KAM-WM records the single-step latent velocity field
\begin{equation}
    \mathbf{V}_{\mathrm{prior}}
    =
    \mathbf{v}_\theta(\mathbf{x}_1,1.0,[o_1,l])
    \in \mathbb{R}^{C\times F\times H\times W},
    \label{eq:vprior}
\end{equation}
which we call a \textbf{Kinematic Affordance Map (KAM)}. The frame axis $F$ is
the model's parallel latent prediction over a fixed temporal window in one
forward pass, with no autoregressive generation. KAM therefore remains in the
video model's latent space: it is not decoded into future frames and is not
interpreted as optical flow or a calibrated 3D displacement. For visualization
and ablations, we separate a channel-axis magnitude from a normalized response:
\begin{equation}
\begin{aligned}
    A_{\mathrm{kam}}(f,h,w)
    &=
    \left\|
        \mathbf{V}_{\mathrm{prior}}(:,f,h,w)
    \right\|_2, \\
    \widehat{\mathbf{V}}_{\mathrm{prior}}(:,f,h,w)
    &=
    \frac{
        \mathbf{V}_{\mathrm{prior}}(:,f,h,w)
    }{
        A_{\mathrm{kam}}(f,h,w)+\epsilon
    } .
\end{aligned}
\label{eq:kam_magnitude}
\end{equation}
The magnitude highlights where the model has a task-conditioned response, while
the normalized response preserves coarse orientation structure in the latent
field. These two views are useful for analysis, but unless noted, the policy
consumes the dense field $\mathbf{V}_{\mathrm{prior}}$ through the Perceiver so
that localization and directional structure are provided jointly.

%===============================================================================
\subsection{Dynamic Token Compression}
\label{subsec:tokens}

The raw KAM tensor is too dense to use directly as a policy condition. We
compress it with a Perceiver cross-attention
bottleneck~\cite{jaegle2021perceiver}: a $1{\times}1{\times}1$ projection and 3D
adaptive average pooling produce a token grid, which is flattened, augmented with
learned positional embeddings, and compressed by $K$ learnable queries,
\begin{equation}
\begin{aligned}
    \mathbf{Z}
    &=
    \mathrm{Flatten}\!\left(
        \mathrm{Pool}\!\left(
            \phi_{1\times1\times1}(\mathbf{V}_{\mathrm{prior}})
        \right)
    \right)
    + \mathbf{E}_{\mathrm{pos}}, \\
    \mathbf{H}_{\mathrm{dyn}}
    &=
    \mathrm{Perceiver}(\mathbf{Q},\mathbf{Z})
    \in \mathbb{R}^{K\times d_k},
\end{aligned}
\label{eq:perceiver_tokens}
\end{equation}
where $\mathbf{Q}$ are the learnable queries and $\mathbf{Z}$ the pooled token
grid. This keeps the prior compact while preserving coarse spatial structure.

%===============================================================================
\subsection{KAM-Conditioned Diffusion Policy}
\label{subsec:policy}

KAM-WM uses a 1D U-Net Diffusion Policy~\cite{chi2023diffusionpolicy} as the
action policy. The KAM prior is already language-conditioned through extraction;
to make the RGB stream instruction-aware, we modulate the observation encoder
$\Psi$ with FiLM~\cite{perez2018film} using the frozen Wan~2.2 text embedding
$\Phi(l)$. The policy conditions on multi-view RGB, proprioception, and the fixed
KAM tokens through a global feature. At control step $\tau$, with proprioceptive
state $s_\tau$ and diffusion step $k$,
\begin{equation}
\begin{aligned}
    \mathbf{g}_\tau
    &=
    \left[
        \Psi(o_\tau,\Phi(l));\,
        \mathrm{vec}(\mathbf{H}_{\mathrm{dyn}});\,
        s_\tau
    \right], \\
    \widetilde{\mathbf{a}}_k
    &=
    \alpha_k \mathbf{a}_{\tau:\tau+H_a}
    +
    \sigma_k \boldsymbol{\epsilon}, \\
    \mathcal{L}_{\mathrm{DP}}
    &=
    \mathbb{E}
    \left[
        \left\|
            \boldsymbol{\epsilon}
            -
            \boldsymbol{\epsilon}_\theta(
                \widetilde{\mathbf{a}}_k,k,\mathbf{g}_\tau
            )
        \right\|_2^2
    \right].
\end{aligned}
\label{eq:policy_objective}
\end{equation}
KAM enters only as a conditioning input, never as an action target. By default
it is extracted once per episode and held fixed, so the backbone runs a single
query per episode while the policy stays closed-loop on RGB and proprioception.

%===============================================================================
\section{Experiments}
\label{sec:experiments}

Our experiments address three questions. First, does a single-step latent
velocity prior improve manipulation in the evaluated low-data settings? Second,
are the observed gains consistent with a first-order directional cue rather than
spatial localization alone? Third, what does this prior cost in inference
latency and trainable parameters? We study the first two on LIBERO and RoboTwin~2.0, with controlled
comparisons against a zero-order mask prior and an analysis of the extraction
timestep, and report efficiency in Section~\ref{subsec:additional_analysis}.

%===============================================================================
\subsection{Experimental Setup}
\label{subsec:setup}

\paragraph{Implementation.}
We use the frozen Wan2.2-TI2V-5B backbone~\cite{wan2025wan}, querying its video
DiT once at the noise endpoint $t{=}1.0$ to read the single-step latent velocity
field $\mathbf{V}_{\mathrm{prior}} \in \mathbb{R}^{48\times21\times50\times68}$
from the first head-camera observation; its text encoder supplies the FiLM
embeddings used by the policy. The backbone is never updated, and we use neither
future frames nor simulator state. Only three components are trained: a Perceiver
bottleneck that compresses $\mathbf{V}_{\mathrm{prior}}$ into $K{=}8$ dynamic
tokens, a per-view ResNet-18 encoder with text-modulated FiLM, and a 1D U-Net
Diffusion Policy with action horizon $H_a{=}16$. KAM is extracted from the head
camera, while the policy consumes both head and wrist-camera RGB together with
proprioception ($14$-DoF dual-arm for RoboTwin~2.0, single-arm for LIBERO).
Full optimizer, schedule, and augmentation settings are in
Appendix~\ref{app:implementation}.

\paragraph{Benchmarks.}
\textbf{LIBERO}~\cite{liu2023libero} comprises 40 tasks across four suites (Spatial,
Object, Goal, Long); we follow the standard 50-demonstration protocol, train one
multi-task policy per suite (500 demonstrations each), and report per-suite
success rates. On LIBERO, DP and OpenVLA results are taken from the OpenVLA
paper under the same benchmark setting. \textbf{RoboTwin~2.0}~\cite{chen2025robotwin} contains
50 contact-rich bimanual tasks; we train each independently from 50
demonstrations and evaluate over 100 rollouts under \textit{Easy} and
\textit{Hard} settings, the latter randomizing object positions, textures, and
lighting. Baseline results are taken from the RoboTwin~2.0 leaderboard; KAM-WM
and prior-type ablations are evaluated under the same 50-demonstration setting
with 100 rollouts per task. Unless otherwise stated, benchmark success rates are
from a single training run and should not be interpreted as multi-seed
estimates. We report representative task results together with the full 50-task
average.

%===============================================================================
\subsection{Results on LIBERO}
\label{subsec:results_libero}

\begin{wraptable}{r}{0.5\textwidth}
  \vspace{-15pt}
  \centering
  \caption{\textbf{LIBERO success rates.} Results are reported in the standard
  50-demonstration LIBERO setting. KAM-WM trains one multi-task policy per suite,
  and other results are taken from ~\cite{kim2024openvla}.}
  \label{tab:libero_results}
  \scriptsize
  \begin{tabular*}{\linewidth}{@{\extracolsep{\fill}} l cccc c @{}}
    \toprule
    \textbf{Method} & \textbf{Spat.} & \textbf{Obj.} & \textbf{Goal} & \textbf{Long} & \textbf{Avg.} \\
    \midrule
    DP~\cite{chi2023diffusionpolicy} & 78.3 & 92.5 & 68.3 & 50.5 & 72.4 \\
    OpenVLA~\cite{kim2024openvla}    & 84.7 & 88.4 & 79.2 & 53.7 & 76.5 \\
    \rowcolor{gray!10}
    \textbf{KAM-WM}                  & \textbf{93.6} & \textbf{96.4} & \textbf{93.6} & \textbf{78.8} & \textbf{90.6} \\
    \bottomrule
  \end{tabular*}
  \vspace{-10pt}
\end{wraptable}

KAM-WM improves LIBERO performance under the standard 50-demo setting. As shown
in Table~\ref{tab:libero_results}, KAM-WM reaches 90.6\% average success,
exceeding the DP and OpenVLA results reported in the OpenVLA paper by 18.2 and
14.1 points, respectively. The gains are most visible
on the Goal and Long suites, where policies must maintain task intent over
longer interactions and recover from intermediate state changes. On the Long
suite, KAM-WM improves from 50.5\% to 78.8\% over DP and from 53.7\% to 78.8\%
over OpenVLA. This pattern is consistent with KAM serving as a task-conditioned
visual prior rather than only a static object localizer.

%===============================================================================
\subsection{Results on RoboTwin~2.0}
\label{subsec:results_robotwin}

KAM-WM achieves the highest 50-task average success among the compared methods under
the 50-demonstration protocol. Table~\ref{tab:robotwin_results} reports
representative tasks together with the full 50-task average. KAM-WM reaches
65.7\% Easy success across 50 tasks. Compared with leaderboard baselines under
the same setting, KAM-WM exceeds the reported DP3 Easy average of 55.2\%.
Because baselines are taken from the leaderboard rather
than reproduced in our codebase, we treat these comparisons as contextual rather
than controlled implementation comparisons. The largest displayed gains appear on tasks involving
directional contact, object placement, or bimanual coordination, such as
\textit{Click Alarmclock}, \textit{Hanging Mug}, \textit{Place Cans Plasticbox},
and \textit{Handover Block}. This pattern is consistent with the intended role
of KAM as a single-step latent velocity field used by the policy.

\begin{table*}[h]
\centering
\caption{\textbf{RoboTwin~2.0 success rates.} Baseline results are taken from
the RoboTwin~2.0 leaderboard; KAM-WM and prior-type ablations are evaluated
under the same 50-demonstration setting with 100 rollouts per task. We report
representative task results together with the full 50-task average.}
\label{tab:robotwin_results}
\scriptsize
\setlength{\tabcolsep}{4.5pt}
\renewcommand{\arraystretch}{1.05}
\begin{tabular}{ll *{9}{c} @{\hskip 4pt} cc}
\toprule
\textbf{Config} & \textbf{Method}
  & \makecell{\textbf{Click}\\\textbf{Alarm.}}
  & \makecell{\textbf{Hang}\\\textbf{Mug}}
  & \makecell{\textbf{Cans}\\\textbf{Plastic.}}
  & \makecell{\textbf{Burger}\\\textbf{Fries}}
  & \makecell{\textbf{Open}\\\textbf{Laptop}}
  & \makecell{\textbf{H'over}\\\textbf{Mic}}
  & \makecell{\textbf{H'over}\\\textbf{Block}}
  & \makecell{\textbf{Press}\\\textbf{Stapler}}
  & \makecell{\textbf{Shake}\\\textbf{Bottle}}
  & \makecell{\textbf{Sel.}\\\textbf{Avg.}}
  & \makecell{\textbf{50-task}\\\textbf{Avg.}} \\
\midrule
\multirow{6}{*}{Easy}
  & ACT~\cite{zhao2023act}
      & 32 & 7  & 16 & 49 & 56 & 85 & 42 & 31 & 74
      & 43.6 & 29.7 \\
  & DP~\cite{chi2023diffusionpolicy}
      & 61 & 8  & 40 & 72 & 49 & 53 & 10 &  6 & 65
      & 40.4 & 28.0 \\
  & RDT~\cite{rdt}
      & 61 & 23 & 6 & 50 & 59 & 90 & 45 & 41 & 74
      & 49.9 & 34.5 \\
  & DP3~\cite{dp3}
      & 77 & 17 & 48 & 72 & 82 & \textbf{100} & 70 & 69 & \textbf{98}
      & 70.3 & 55.2 \\
  & $\pi_0$ (FT)~\cite{black2024pi0}
      & 63 & 11 & 34 & 80 & \textbf{85} & 98 & 45 & 62 & 97
      & 63.9 & 46.4 \\
  \rowcolor{gray!10}
  & \textbf{KAM-WM}
      & \textbf{96} & \textbf{47} & \textbf{87} & \textbf{96}
      & 84 & 95 & \textbf{85} & \textbf{71} & 93
      & \textbf{83.8} & \textbf{65.7} \\
\midrule
\multirow{6}{*}{Hard}
  & ACT~\cite{zhao2023act}
      & 4 & 0 & 0 & 0 & 0 & 0 & 0 &  6 & 10
      & 2.2 & 1.7 \\
  & DP~\cite{chi2023diffusionpolicy}
      & 5 & 0 & 0 & 0 & 0 & 0 & 0 &  0 &  8
      & 1.4 & 0.6 \\
  & RDT~\cite{rdt}
      & 12 & \textbf{16} & 5 & 27 & 32 & 31 & 14 & 24 & 45
      & 22.9 & 13.7 \\
  & DP3~\cite{dp3}
      & 14 & 1 & 3 & 18 & 7 & 3 & 0 & 3 & 19
      & 7.6 & 5.0 \\
  & $\pi_0$ (FT)~\cite{black2024pi0}
      & 11 & 3 & 2 & 4 & 46 & 13 & 8 & 29 & \textbf{60}
      & 19.6 & 16.3 \\
  \rowcolor{gray!10}
  & \textbf{KAM-WM}
      & \textbf{45} & 9 & \textbf{25} & \textbf{40}
      & \textbf{69} & \textbf{80} & \textbf{35} & \textbf{31} & 59
      & \textbf{43.7} & \textbf{22.4} \\
\bottomrule
\end{tabular}
\end{table*}

The Hard setting remains difficult for all methods. KAM-WM reaches 22.4\%
Hard success across 50 tasks, compared with leaderboard results of 16.3\% for
$\pi_0$, 13.7\% for RDT, 5.0\% for DP3, 1.7\% for ACT, and 0.6\% for DP.
The displayed tasks illustrate per-task behavior, while the 50-task
average is the primary aggregate. KAM-WM tends to gain most on tasks
with asymmetric approach geometry, where identifying the interaction side is
important for success.

%===============================================================================
\subsection{Ablation Studies}
\label{subsec:analysis}

We run three diagnostic studies: 1) the conditioning prior (KAM vs.\ a
zero-order mask), 2) the KAM extraction timestep, and 3) behavior in a
lower-data regime. All three hold the policy architecture and protocol fixed and
are diagnostic rather than full-benchmark sweeps; additional prior comparisons
on RoboTwin tasks are in Appendix~\ref{app:extended_svpil}.

\begin{figure*}[t]
    \centering
    \includegraphics[width=\textwidth]{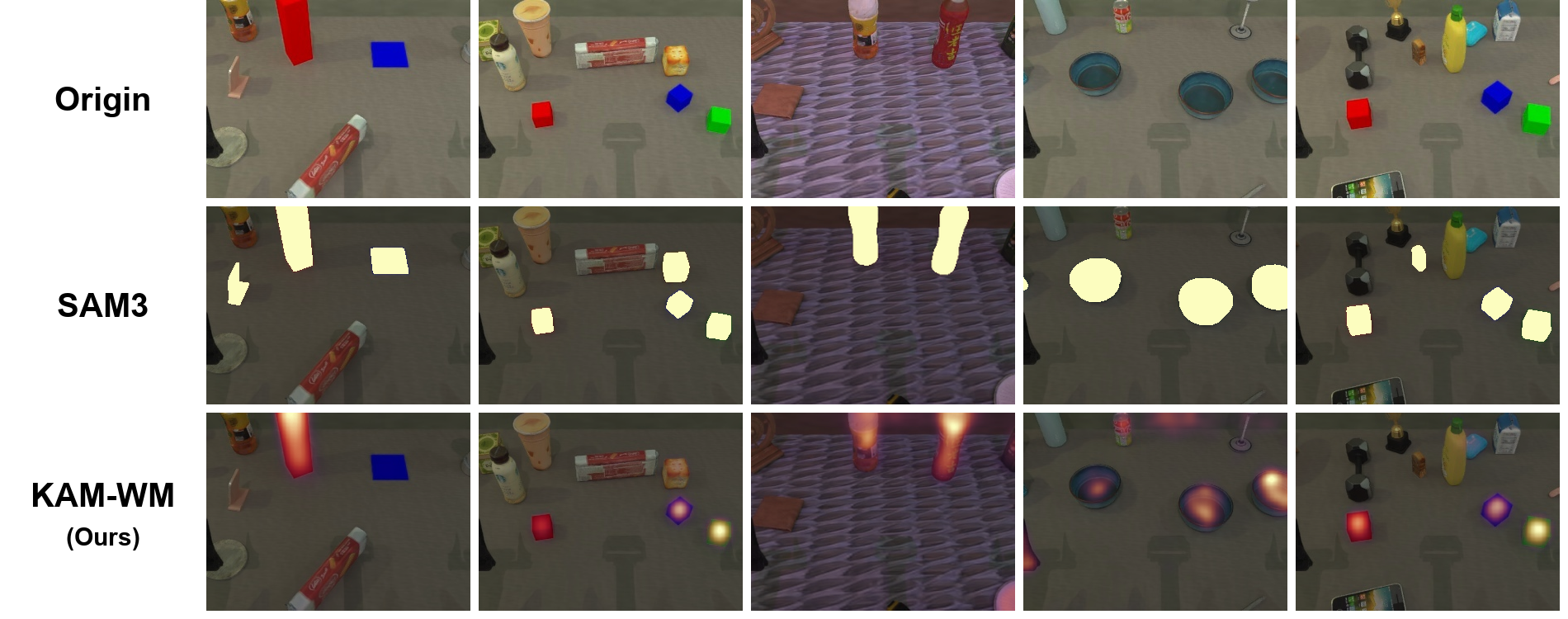}
    \caption{\textbf{Qualitative comparison of mask and KAM priors.}
    Mask priors localize target regions but remain zero-order spatial prior.
    KAM highlights more selective interaction regions and adds coarse
    orientation prior from the latent velocity field under the same policy
    interface.}
    \label{fig:kam_vs_mask}
\end{figure*}

\paragraph{Prior type.}
We compare KAM with a zero-order mask prior while keeping the policy architecture
fixed. The mask baseline follows the spatial visual prompt setup from our prior
SVP work~\cite{tang2026svp}: it uses SAM 3~\cite{carion2025sam3} masks conditioned
on the task instruction, rendered at the KAM resolution and passed through the
same Perceiver bottleneck, so only the conditioning signal changes from a
language-conditioned mask to KAM. As shown in
Table~\ref{tab:ablation_studies}, KAM raises the Easy average from 67.8\% to
77.0\% and the Hard average from 24.0\% to 28.6\%, with the largest Easy gains on
\textit{Hanging Mug} and \textit{Place Empty Cup}, where approach direction and
post-contact motion matter. KAM does not always win: on \textit{Click Bell} it
improves Easy but drops from 34\% to 18\% on Hard, and a similar drop appears on
\textit{Place Shoe} (22\% vs 10\% Hard). A plausible explanation is that the
latent velocity direction can be more sensitive to texture and lighting drift
than mask-based localization when precise point contact is required.
Figure~\ref{fig:kam_vs_mask} shows qualitative examples.

%===============================================================================

\begin{wraptable}{r}{0.52\textwidth}
  \vspace{-8pt}
  \centering
  \caption{\textbf{Prior-type ablation on RoboTwin~2.0.} KAM vs.\ a zero-order
  mask prior under the same policy interface.}
  \label{tab:ablation_studies}
  \vspace{1mm}
  \scriptsize
  \setlength{\tabcolsep}{2.4pt}
  \renewcommand{\arraystretch}{0.90}
  \resizebox{\linewidth}{!}{%
  \begin{tabular}{@{}lcccc@{}}
  \toprule
  \multirow{2}{*}{\textbf{Task}} &
  \multicolumn{2}{c}{\textbf{Easy}} &
  \multicolumn{2}{c}{\textbf{Hard}} \\
  \cmidrule(lr){2-3}\cmidrule(l){4-5}
  & +Mask & \textbf{+KAM} & +Mask & \textbf{+KAM} \\
  \midrule
  Adjust Bottle         & 96 & \textbf{100} & 76 & \textbf{87} \\
  Click Bell            & 83 & \textbf{94}  & \textbf{34} & 18 \\
  Hanging Mug           & 24 & \textbf{47}  & \textbf{9}  & \textbf{9} \\
  Lift Pot              & \textbf{90} & 89 & 13 & \textbf{17} \\
  Place Can Basket      & \textbf{42} & 40 & 15 & \textbf{26} \\
  Place Cans Plasticbox & \textbf{88} & 87 & 4  & \textbf{25} \\
  Place Empty Cup       & 49 & \textbf{79}  & 4  & \textbf{6} \\
  Place Shoe            & 45 & \textbf{64}  & \textbf{22} & 10 \\
  Shake Bottle          & \textbf{93} & \textbf{93} & 39 & \textbf{59} \\
  \midrule
  \rowcolor{gray!10}
  \textbf{Average}      & 67.8 & \textbf{77.0} & 24.0 & \textbf{28.6} \\
  \bottomrule
  \end{tabular}%
  }
  \vspace{-10pt}
\end{wraptable}

\paragraph{Extraction timestep.}
We study where along the Flow Matching trajectory to extract the field. Any
$t<1.0$ requires denoising from the noise endpoint to the target step,
reintroducing multi-step cost, so $t{=}1.0$ is the only strictly rollout-free
operating point.
Table~\ref{tab:ablation_timestep} shows the best timestep is task dependent:
$t{=}0.0$ or $t{=}0.5$ can help some tasks, so $t{=}1.0$ is not oracle-best. We
adopt $t{=}1.0$ because it gives a competitive prior from one fixed query
without partial video generation.

\paragraph{Low-data regime.}
To probe whether first-order priors help as demonstrations grow scarcer, we
compare the Diffusion Policy backbone, the same backbone with a zero-order mask
prior, and DP+KAM-WM under 50-demo and 20-demo Easy settings, using the same
checkpoint-selection rule for all methods (best of epochs \{100, 300, 600\}). The
six tasks span direction-sensitive placement, point contact, geometry-sensitive
lifting, repeated contact, and localization-dominant placement.
Table~\ref{tab:lowdata_mask} shows the mask prior is neutral at 50 demos and
mixed at 20 demos on this subset, whereas 20-demo KAM is uniformly non-negative
relative to the backbone and raises the subset average from 45.0\% to 66.7\%.
We therefore view KAM as promising in lower-data regimes, while leaving
full-suite 20-demo evaluation to future work.

\begin{table*}[t]
\centering
\begin{minipage}[t]{0.45\textwidth}
\vspace{0pt}
\centering
\setcounter{table}{3}
\caption{\textbf{Extraction-timestep ablation.} Later timesteps can improve some
tasks, but $t{=}1.0$ is the only rollout-free setting.}
\label{tab:ablation_timestep}
\vspace{1mm}
\scriptsize
\setlength{\tabcolsep}{3.0pt}
\renewcommand{\arraystretch}{1.15}
\resizebox{\linewidth}{!}{%
\begin{tabular}{@{}lcccccc@{}}
\toprule
\multirow{2}{*}{\textbf{Task}} &
\multicolumn{2}{c}{$t{=}0.0$} &
\multicolumn{2}{c}{$t{=}0.5$} &
\multicolumn{2}{c}{$t{=}1.0$} \\
\cmidrule(lr){2-3}\cmidrule(lr){4-5}\cmidrule(l){6-7}
& Easy & Hard & Easy & Hard & Easy & Hard \\
\midrule
Beat Hammer      & \textbf{91} & \textbf{17} & 88 & 12 & 88 & 12 \\
Cans Plasticbox  & 60 & 26 & 85 & \textbf{27} & \textbf{87} & 25 \\
Hanging Mug      & \textbf{59} & \textbf{11} & 44 & 7  & 47 & 9  \\
Handover Block   & 90 & \textbf{42} & \textbf{91} & 33 & 56 & 35 \\
\midrule
\rowcolor{gray!10}
\textbf{Average} & 75 & \textbf{24} & \textbf{77} & 20 & 70 & 20 \\
\bottomrule
\end{tabular}%
}
\end{minipage}\hfill
\begin{minipage}[t]{0.52\textwidth}
\vspace{0pt}
\centering
\setcounter{table}{4}
\caption{\textbf{Targeted low-data diagnostic.} Easy-setting success rates on
six representative RoboTwin~2.0 tasks.}
\label{tab:lowdata_mask}
\vspace{1mm}
\scriptsize
\setlength{\tabcolsep}{2.8pt}
\renewcommand{\arraystretch}{0.95}
\resizebox{\linewidth}{!}{%
\begin{tabular}{@{}lcccccc@{}}
\toprule
\multirow{2}{*}{\textbf{Task}} &
\multicolumn{3}{c}{\textbf{50-demo}} &
\multicolumn{3}{c}{\textbf{20-demo}} \\
\cmidrule(lr){2-4}\cmidrule(l){5-7}
& \textbf{DP} & \textbf{+Mask} & \textbf{+KAM} & \textbf{DP} & \textbf{+Mask} & \textbf{+KAM} \\
\midrule
Adjust Bottle   & 100 & 100 & 100 & 60 & 50 & 90 \\
Lift Pot        & 90  & 90  & 89  & 35 & 45 & 55 \\
Click Bell      & 95  & 95  & 94  & 40 & 30 & 75 \\
Shake Bottle    & 90  & 85  & 93  & 75 & 70 & 75 \\
Open Laptop     & 90  & 85  & 84  & 55 & 55 & 75 \\
Place Empty Cup & 25  & 35  & 79  & 5  & 10 & 30 \\
\midrule
\rowcolor{gray!10}
\textbf{Average} & 81.7 & 81.7 & 89.8 & 45.0 & 43.3 & 66.7 \\
\bottomrule
\end{tabular}%
}
\end{minipage}
\setcounter{table}{5}
\end{table*}

%===============================================================================
\subsection{Efficiency Analysis}
\label{subsec:additional_analysis}

\paragraph{Inference efficiency.}
KAM-WM avoids multi-step video rollout during policy inference. With 10
denoising steps, the action-policy latency is \textbf{99\,ms} per action chunk,
compared with \textbf{380\,ms} for our fine-tuned $\pi_0$ implementation on the
same evaluation workstation. This per-chunk number excludes the one-time KAM
extraction, which costs \textbf{895\,ms} once per episode and is then amortized
over the rollout. Perceiver compression adds less than \textbf{1\,ms}, and
cached KAM tensors of about \textbf{6.0\,MB} per episode keep steady-state
inference dominated by the policy.

\paragraph{Parameter efficiency.}
KAM-WM keeps the 5B video backbone frozen and precomputes the KAM tensors once
per episode, so it is not in the policy-training loop. The trainable components
are the per-view ResNet-18 encoder, the $93.4$M 1D U-Net policy, and the $2.1$M
Perceiver bottleneck, totaling roughly $120$M excluding frozen Wan parameters.
The KAM-specific trainable module is the $2.1$M Perceiver. KAM-WM reaches
65.7\% Easy 50-task success, compared contextually with the 28.0\% DP
leaderboard average.
Compared with the reported $\pi_0$ FT baseline, KAM-WM trains roughly
$25\times$ fewer parameters while reaching 65.7\% versus 46.4\% Easy success.
The prior-type ablation in Table~\ref{tab:ablation_studies} further keeps the
architecture fixed and swaps only the conditioning signal, reducing the chance
that the gains are explained by added policy capacity alone.

%===============================================================================
\section{Limitations, Discussion, and Conclusion}
\label{sec:limitations}
\label{sec:conclusion}

\paragraph{Limitations and Discussion.}
KAM-WM has five main limitations. First, it extracts KAM once per episode from
the first head-camera frame, which keeps inference lightweight but lets the prior
go stale under long horizons, major scene changes, or occlusion; refreshing it
at task-critical moments is a natural extension. Second, the single-step
velocity field is tied to the Flow Matching parameterization and should be
re-examined for other video generators. Third, KAM is a coarse visual-motion
prior rather than an explicit 3D plan, so precise contact or insertion may need
tighter geometric, tactile, or multi-view feedback. Fourth, our evaluation is
simulation-only: Wan2.2 rollouts from real robot images provide only a
qualitative sanity check, not real-world validation. Fifth, we report single-run
success rates rather than multi-seed confidence intervals, leaving training
variance in the low-data regime to future work.

\paragraph{Conclusion.}
We introduced \textbf{KAM-WM}, which reuses the single-step latent velocity
field of a frozen Flow Matching image-to-video backbone as a first-order visual
prior for low-data robot manipulation. Rather than rolling out future frames or
fine-tuning the backbone, KAM-WM reads a task-conditioned motion cue in a single
query and lets a lightweight policy learn to use it. Across LIBERO and
RoboTwin~2.0 it improves over reported DP/OpenVLA and leaderboard baselines in
contextual comparisons, while training over $25\times$ fewer parameters than the
fine-tuned $\pi_0$ comparison. More broadly, our results suggest that
some directional structure encoded by a frozen video model is accessible without
test-time future generation, making such models a useful source of first-order
visual priors in the evaluated manipulation settings.

%===============================================================================
% No \bibliographystyle is required because the CoRL style is automatically used.
\clearpage
\bibliography{reference_arxiv}

\clearpage
\appendix

\section*{Appendix Overview}
This appendix provides implementation details, an auxiliary real-world
feasibility note, extended prior-type comparisons, and the complete
RoboTwin~2.0 results table:
\begin{itemize}[itemsep=2pt, topsep=3pt, parsep=0pt, partopsep=0pt]
    \item Appendix~\ref{app:implementation}: Backbone usage, policy architecture, optimization, and evaluation protocol.
    \item Appendix~\ref{app:realworld}: Auxiliary real-world feasibility note and representative outcomes.
    \item Appendix~\ref{app:extended_svpil}: Extended comparison with zero-order mask priors.
    \item Appendix~\ref{app:full_robotwin}: Complete 50-task RoboTwin~2.0 results.
\end{itemize}

%===============================================================================
\section{Implementation Details}
\label{app:implementation}

\paragraph{Latent world model and prior extraction.}
We use Wan~2.2 5B~\cite{wan2025wan} as a frozen image-to-video backbone. Given
the first head-camera observation of an episode
$o_1 \in \mathbb{R}^{3 \times 480 \times 640}$ and the language instruction $l$,
we query the model once at the high-noise endpoint $t{=}1.0$. We record the
resulting latent velocity field
\[
    \mathbf{V}_{\mathrm{prior}} \in \mathbb{R}^{48 \times 21 \times 50 \times 68}
\]
after a single forward pass through the DiT stack. No iterative denoising or
future-frame generation is performed. The world-model backbone remains frozen
during both training and evaluation.

\paragraph{KAM-WM Perceiver bottleneck.}
The Perceiver bottleneck compresses the dense KAM tensor into a small set of
dynamic tokens. A $1{\times}1{\times}1$ Conv3D projection maps the channel
dimension from 48 to 256, followed by GELU activation and 3D adaptive average
pooling to $(7,10,17)$. This yields $M'=1{,}190$ tokens. We add a learned
positional embedding
$\mathbf{E}_{\mathrm{pos}} \in \mathbb{R}^{1190 \times 256}$ initialized with a
truncated normal distribution ($\sigma=0.02$). $K=8$ learnable latent queries
attend to these tokens through one pre-norm multi-head cross-attention layer
with 8 heads and hidden size 256, followed by a pre-norm feed-forward network
with hidden size $4 \times 256$. The bottleneck contains approximately 2.1M
trainable parameters.

\paragraph{Visual encoder with text-modulated FiLM.}
The multi-view visual encoder $\Psi$ is a ResNet-18 instantiated independently
for each camera view. After each residual block, feature maps are modulated by
FiLM layers~\cite{perez2018film}. The FiLM scale and shift parameters are
predicted from the frozen Wan~2.2 text embedding $\Phi(l)$ using a small MLP.
The ResNet and FiLM heads are trained jointly with the Perceiver and diffusion
policy, while the Wan text encoder remains frozen.

\paragraph{1D U-Net Diffusion Policy.}
The downstream policy follows a standard 1D U-Net Diffusion Policy
architecture~\cite{chi2023diffusionpolicy}. It uses four downsampling stages
with channel widths 256--512--1024--1024. The global condition concatenates
language-modulated visual features, flattened KAM tokens
($K \times d_k = 2{,}048$ dimensions), and proprioceptive features. For
RoboTwin~2.0, proprioception includes the 14-DoF dual-arm state and
end-effector poses. For LIBERO, it uses the corresponding single-arm state. The
action chunk horizon is $H_a=16$. We use a DDPM scheduler with 100 denoising
steps during training and 10 steps during inference.

\paragraph{Training configuration.}
For RoboTwin~2.0, each task is trained independently. For LIBERO, we train a
single multi-task policy for each suite. All policies are trained for
3{,}000 gradient steps with batch size 64
on a single high-end GPU. The Perceiver, visual encoder, FiLM heads, and U-Net
policy are optimized with AdamW using learning rate $1\times10^{-4}$, weight
decay $10^{-4}$, and cosine learning-rate decay. Training each model takes
approximately 2--3 GPU hours. We apply random color jitter and horizontal flips
to each camera view.

\paragraph{Prior caching during training.}
For the default setting used in the paper, KAM is extracted from the first
observation of each demonstration episode and reused for all training windows
from that episode. We therefore precompute and cache the KAM tensor once per
episode. During training, the frozen 5B image-to-video backbone is removed from
the hot path; the training loop reads cached priors and only updates the Perceiver,
visual encoder, and diffusion policy.

\paragraph{Evaluation protocol.}
Each RoboTwin~2.0 task is evaluated over 100 rollouts. Unless otherwise stated,
the reported results are from a single training run per method. The Easy setting follows
RoboTwin's Clean setting, with a clean background and no distractors, while the
Hard setting follows RoboTwin's Random setting, with complex backgrounds,
challenging lighting, and 10 distractors in each scene. At test time, KAM is extracted once from the
first observation of an episode and held fixed during closed-loop execution. The
reported per-chunk policy latency excludes this one-time prior extraction, which
takes approximately 895\,ms on the evaluation workstation GPU. The policy
latency is measured during closed-loop action prediction with 10 denoising
steps and excludes simulator stepping, rendering, and environment reset time.

\section{Real-World Deployment Study}
\label{app:realworld}

\newcommand{\NUMSTEPS}{20{,}000}

The main experiments in this paper use a Diffusion Policy (1D U-Net) backbone. This
section instead deploys KAM on a $\pi_0$ backbone to test whether KAM behaves as a
plug-and-play conditioning prior that transfers across policy backbones, rather
than as a component tied to a single architecture. We adapt the KAM prior to the
$\pi_0$ conditioning interface and evaluate it on a physical single-arm platform. We
treat this as a feasibility study under a limited task and demonstration budget.
It is not a controlled comparison, and we do not pair it with reproduced
real-world baselines.

\paragraph{Hardware platform and sensing.}
The platform is a single-arm AgileX PiPER with six degrees of freedom. Three RGB
cameras at $640\times480$ resolution provide visual input. A top camera mounted
above the workspace supplies the view used for KAM extraction. A wrist camera on
the end-effector provides close-range feedback near contact. A third camera,
mounted below the top camera, adds an external view of the workspace.
All three views feed the real-world policy, whereas the simulation policy uses
the head and wrist camera views. KAM is still extracted from the top/head view
only, so the prior remains a single-view, single-query input as in simulation.

\paragraph{Deployment pipeline.}
All three streams are resized to the policy input resolution used during
training. The policy then runs in closed loop on RGB observations and
proprioceptive state, and the predicted action chunks
are executed on the physical controller. Relative to the simulated setup, real deployment replaces rendered observations
with real camera streams and executes predicted action chunks on the physical
controller. In addition, because this study uses a $\pi_0$ backbone rather than
the Diffusion Policy backbone used in the main simulation experiments, we adapt
the KAM injection mechanism to the $\pi_0$ conditioning interface. No
real-world-specific prior is introduced. We keep the same single-query KAM prior
used in simulation, but its injection into the policy is adapted to the $\pi_0$
conditioning interface rather than the Diffusion Policy interface used for the
main results. Concretely, the same Perceiver
bottleneck compresses the KAM tensor into 8 latent tokens, which are
projected to the backbone hidden width and appended to the $\pi_0$ prefix sequence
alongside the image and language tokens, so the downstream state and action
tokens attend to them as full-attention prefix context.

\begin{figure}[ht]
    \centering
    \begin{subfigure}{\linewidth}
        \centering
        \includegraphics[width=0.95\linewidth]{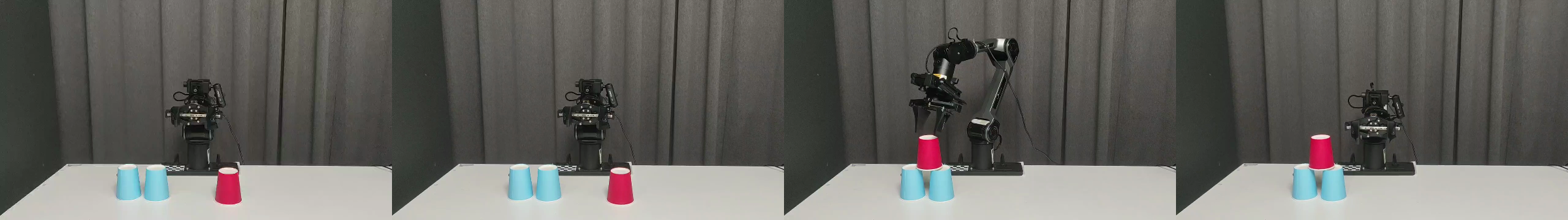}
        \caption{Pick up the red cup and stack it among the blue cups}
    \end{subfigure}
    \vspace{3mm}
    \begin{subfigure}{\linewidth}
        \centering
        \includegraphics[width=0.95\linewidth]{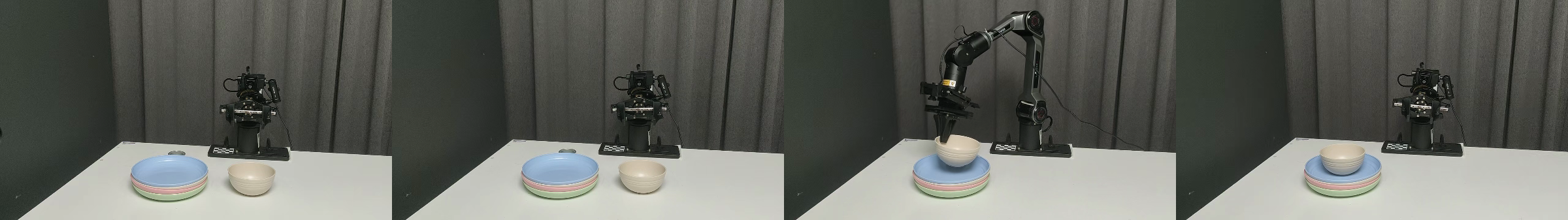}
        \caption{Pick up the bowl and place it on the plates}
    \end{subfigure}
    \vspace{3mm}
    \begin{subfigure}{\linewidth}
        \centering
        \includegraphics[width=0.95\linewidth]{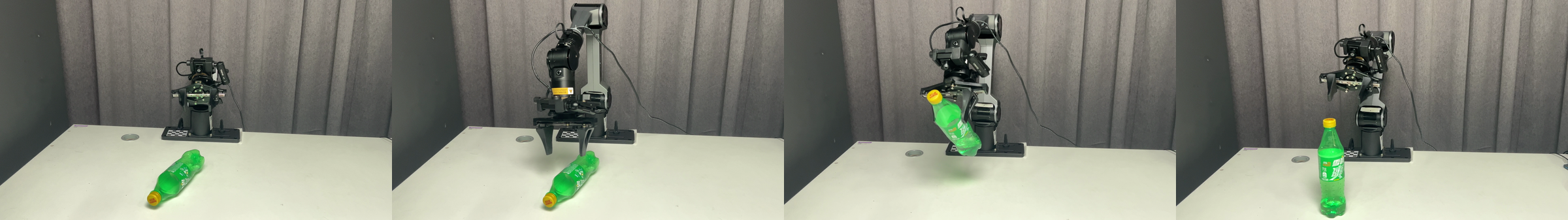}
        \caption{Pick up the bottle and place it upright}
    \end{subfigure}
    \vspace{3mm}
    \begin{subfigure}{\linewidth}
        \centering
        \includegraphics[width=0.95\linewidth]{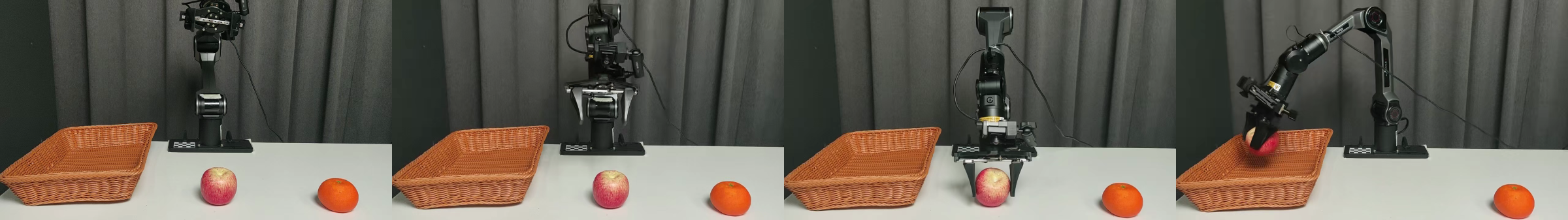}
        \caption{Pick up the apple and place it in the basket}
    \end{subfigure}
    \caption{\textbf{Real-world setup and representative execution sequences.}
    Key frames from four real-world tasks:
    (a) cup stacking, (b) bowl-to-plates, (c) place bottle, and
    (d) apple-to-basket.}
    \label{fig:realworld_setups}
\end{figure}

\paragraph{Tasks and training protocol.}
We consider four manipulation tasks: cup stacking, bowl-to-plates, place
bottle, and apple-to-basket (Figure~\ref{fig:realworld_setups}). For each task, we collect 30
teleoperated demonstrations, hold out 2 for validation, and train a separate
policy for \NUMSTEPS{} gradient steps under a 28:2 train/validation split. The
held-out demonstrations are used for checkpoint selection only.

\paragraph{Evaluation protocol.}
Each policy is evaluated over 20 physical rollouts, with a task-dependent
maximum episode length set by the execution budget of the task. A rollout is
scored as successful when the robot completes the task objective: stacking the
red cup among the blue cups, placing the bowl onto the plates, placing the
bottle upright, or placing the apple in the basket. We report the number of
successful rollouts out of 20 for each task in
Table~\ref{tab:realworld_results}.

\begin{table}[H]
\caption{\textbf{KAM-WM ($\pi_0$ backbone) real-world deployment results.} Each policy is trained on 30
teleoperated demonstrations and evaluated over 20 physical rollouts.}
\label{tab:realworld_results}
\centering
\small
\setlength{\tabcolsep}{6pt}
\begin{tabular}{l c c c}
\toprule
\textbf{Task} & \textbf{Expert Demos} & \textbf{Train Steps} & \textbf{Success Rate} \\
\midrule
\textit{Cup stacking}   & 30 & \NUMSTEPS{} & 9/20 (45\%) \\
\textit{Bowl-to-plates} & 30 & \NUMSTEPS{} & 13/20 (65\%) \\
\textit{Place bottle}   & 30 & \NUMSTEPS{} & 8/20 (40\%) \\
\textit{Apple-to-basket} & 30 & \NUMSTEPS{} & 12/20 (60\%) \\
\bottomrule
\end{tabular}
\end{table}

\paragraph{Outcomes and failure modes.}
KAM-WM ($\pi_0$ backbone) succeeds in 9, 13, 8, and 12 of 20 rollouts on cup stacking,
bowl-to-plates, place bottle, and apple-to-basket. The policy completes the
task sequence under real sensing and actuation across all four tasks, supporting
the feasibility of the single-query KAM interface on hardware. We do not draw
quantitative conclusions from four tasks at this scale. Remaining failures are
primarily due to grasp misalignment, degraded
wrist-camera feedback near contact, and placement instability at the end of the
trajectory. A controlled real-world ablation that keeps the backbone fixed
while toggling the KAM prior is left for future work.

\section{Additional Comparison with Zero-Order Masks}
\label{app:extended_svpil}

This appendix extends the prior-type ablation in the main paper with a broader
task-level comparison between KAM and zero-order mask priors. The mask-prior
results come from the spatial visual prompt setting in our prior
SVP work~\cite{tang2026svp}. As in the main text, the policy architecture, training
protocol, and evaluation procedure are held fixed, and only the conditioning
signal is changed. The mask baseline uses instruction-conditioned segmentation
masks passed through the same Perceiver interface, so the comparison isolates
the effect of replacing a zero-order spatial prior with the KAM latent velocity
prior.

Table~\ref{tab:appendix_svpil_comparison} reports nine representative
RoboTwin~2.0 tasks under both Easy and Hard settings. The results are broadly
consistent with the trend in the main paper: KAM improves the average over the
mask prior in both Easy and Hard settings, with especially large gains on tasks
such as \textit{Hanging Mug} and \textit{Place Empty Cup}, where initial
approach direction and post-contact motion matter. At the same time, KAM is not
uniformly better on every task, which is also consistent with the discussion in
the main text that latent directional cues can be more sensitive than mask-only
localization in some visually challenging contact scenarios.

\begin{table}[ht]
\caption{\textbf{Extended prior-type comparison on RoboTwin~2.0.} Success rates
(\%) are evaluated over 100 rollouts per task. DP, ACT, and $\pi_0$ rows are
included as contextual references; the controlled comparison is between
\textit{Mask} and \textit{KAM-WM}, which use the same policy interface.}
\label{tab:appendix_svpil_comparison}
\centering
\small
\setlength{\tabcolsep}{3.5pt}
\begin{tabular}{l ccccc ccccc}
\toprule
\multirow{2}{*}{\textbf{Task}} &
  \multicolumn{5}{c}{\textbf{Easy}} &
  \multicolumn{5}{c}{\textbf{Hard}} \\
\cmidrule(lr){2-6}\cmidrule(lr){7-11}
& DP & ACT & $\pi_0$ & Mask & \textbf{KAM-WM}
& DP & ACT & $\pi_0$ & Mask & \textbf{KAM-WM} \\
\midrule
Adjust Bottle
  & 97 & 97 & 90 & 96 & \textbf{100}
  & 66 & 23 & 56 & 76 & \textbf{87} \\
Click Bell
  & 54 & 58 & 44 & 83 & \textbf{94}
  & 4  & 3  & 3  & \textbf{34} & 18 \\
Hanging Mug
  & 8  & 7  & 11 & 24 & \textbf{47}
  & 6  & 0  & 3  & \textbf{9}  & \textbf{9}  \\
Lift Pot
  & 39 & 88 & 84 & \textbf{90} & 89
  & 9  & 0  & \textbf{36} & 13 & 17 \\
Place Can Basket
  & 18 & 1  & \textbf{41} & \textbf{42} & 40
  & 8  & 0  & 5  & 15 & \textbf{26} \\
Place Cans Plasticbox
  & 40 & 16 & 34 & \textbf{88} & 87
  & 4  & 0  & 2  & 4  & \textbf{25} \\
Place Empty Cup
  & 37 & 61 & 37 & 49 & \textbf{79}
  & 2  & 0  & \textbf{11} & 4  & 6  \\
Place Shoe
  & 23 & 5  & 28 & 45 & \textbf{64}
  & 7  & 0  & 6  & \textbf{22} & 10 \\
Shake Bottle
  & 65 & 74 & \textbf{97} & 93 & 93
  & 28 & 10 & \textbf{60} & 39 & 59 \\
\midrule
\textbf{Average}
  & 42.3 & 45.2 & 51.8 & 67.8 & \textbf{77.0}
  & 14.9 & 4.0  & 20.7 & 24.0 & \textbf{28.6} \\
\bottomrule
\end{tabular}
\end{table}

%===============================================================================
\section{Full Evaluation Results on RoboTwin~2.0}
\label{app:full_robotwin}

Table~\ref{tab:full_results_combined} reports the complete per-task evaluation
on all 50 RoboTwin~2.0 tasks under Easy and Hard settings.

The per-task results suggest that KAM-WM performs especially well on tasks where
localized contact and approach direction are important, including
\textit{Click Alarmclock}, \textit{Hanging Mug}, \textit{Place Burger Fries},
and \textit{Handover Block}. These results are consistent with the main paper's
hypothesis that the latent velocity field provides useful information beyond
object-level localization.

KAM-WM is less consistently advantageous on tasks where the interaction is
kinematically simple or where success depends more on semantic disambiguation
and precise execution, such as \textit{Shake Bottle} or \textit{Grab Roller}.
This pattern supports the interpretation of KAM as a coarse first-order visual
prior rather than a complete action representation.

{
\setlength{\LTcapwidth}{\textwidth}
\setlength{\tabcolsep}{0.2pt}
\begin{longtable}{@{}>{\footnotesize\raggedright\arraybackslash}p{34mm}
  *{12}{>{\small\centering\arraybackslash}p{\dimexpr(\textwidth-34mm-24\tabcolsep)/12\relax}}@{}}
\caption{\textbf{Complete RoboTwin~2.0 results over 50 tasks under Easy and
Hard settings.} Success rates (\%) are computed over 100 rollouts per task.
Baseline results are taken from the RoboTwin~2.0 leaderboard; KAM (short for KAM-WM) is
evaluated under the same 50-demonstration setting.}
\label{tab:full_results_combined} \\
\toprule
\multirow{2}{*}{\small\textbf{Task}} &
  \multicolumn{6}{c}{\small\textbf{Easy}} &
  \multicolumn{6}{c}{\small\textbf{Hard}} \\
\cmidrule(lr){2-7}\cmidrule(l){8-13}
& \small\textbf{ACT} & \small\textbf{DP} & \small\textbf{RDT} & \small\textbf{DP3} & \small\textbf{$\pi_0$} & \small\textbf{KAM}
& \small\textbf{ACT} & \small\textbf{DP} & \small\textbf{RDT} & \small\textbf{DP3} & \small\textbf{$\pi_0$} & \small\textbf{KAM} \\
\midrule
\endfirsthead

\multicolumn{13}{c}{\small\textit{Table~\ref{tab:full_results_combined} continued}} \\
\toprule
\multirow{2}{*}{\small\textbf{Task}} &
  \multicolumn{6}{c}{\small\textbf{Easy}} &
  \multicolumn{6}{c}{\small\textbf{Hard}} \\
\cmidrule(lr){2-7}\cmidrule(l){8-13}
& \small\textbf{ACT} & \small\textbf{DP} & \small\textbf{RDT} & \small\textbf{DP3} & \small\textbf{$\pi_0$} & \small\textbf{KAM}
& \small\textbf{ACT} & \small\textbf{DP} & \small\textbf{RDT} & \small\textbf{DP3} & \small\textbf{$\pi_0$} & \small\textbf{KAM} \\
\midrule
\endhead

\midrule
\multicolumn{13}{r}{\small\textit{Continued on next page}} \\
\endfoot

\bottomrule
\endlastfoot

Adjust Bottle             & 97.0 & 97.0 & 81.0 & 99.0 & 90.0 & \textbf{100.0} & 23.0 & 0.0 & 75.0 & 3.0 & 56.0 & \textbf{87.0} \\
Beat Block Hammer         & 56.0 & 42.0 & 77.0 & 72.0 & 43.0 & \textbf{88.0} & 3.0 & 0.0 & \textbf{37.0} & 8.0 & 21.0 & 12.0 \\
Blocks Ranking RGB        & 1.0 & 0.0 & 3.0 & 3.0 & 19.0 & \textbf{28.0} & 0.0 & 0.0 & 0.0 & 0.0 & \textbf{5.0} & 3.0 \\
Blocks Ranking Size       & 0.0 & 1.0 & 0.0 & 2.0 & 7.0 & \textbf{20.0} & 0.0 & 0.0 & 0.0 & 0.0 & \textbf{1.0} & \textbf{1.0} \\
Click Alarmclock          & 32.0 & 61.0 & 61.0 & 77.0 & 63.0 & \textbf{96.0} & 4.0 & 5.0 & 12.0 & 14.0 & 11.0 & \textbf{45.0} \\
Click Bell                & 58.0 & 54.0 & 80.0 & 90.0 & 44.0 & \textbf{94.0} & 3.0 & 0.0 & 9.0 & 0.0 & 3.0 & \textbf{18.0} \\
Dump Bin Bigbin           & 68.0 & 49.0 & 64.0 & \textbf{85.0} & 83.0 & 72.0 & 1.0 & 0.0 & 32.0 & \textbf{53.0} & 24.0 & 47.0 \\
Grab Roller               & 94.0 & \textbf{98.0} & 74.0 & \textbf{98.0} & 96.0 & 97.0 & 25.0 & 0.0 & 43.0 & 2.0 & \textbf{80.0} & \textbf{80.0} \\
Handover Block            & 42.0 & 10.0 & 45.0 & 70.0 & 45.0 & \textbf{85.0} & 0.0 & 0.0 & 14.0 & 0.0 & 8.0 & \textbf{35.0} \\
Handover Mic              & 85.0 & 53.0 & 90.0 & \textbf{100.0} & 98.0 & 95.0 & 0.0 & 0.0 & 31.0 & 3.0 & 13.0 & \textbf{80.0} \\
Hanging Mug               & 7.0 & 8.0 & 23.0 & 17.0 & 11.0 & \textbf{47.0} & 0.0 & 0.0 & \textbf{16.0} & 1.0 & 3.0 & 9.0 \\
Lift Pot                  & 88.0 & 39.0 & 72.0 & \textbf{97.0} & 84.0 & 89.0 & 0.0 & 0.0 & 9.0 & 0.0 & \textbf{36.0} & 17.0 \\
Move Can Pot              & 22.0 & 39.0 & 25.0 & 70.0 & 58.0 & \textbf{87.0} & 4.0 & 0.0 & 12.0 & 6.0 & 21.0 & \textbf{28.0} \\
Move Pillbottle Pad       & 0.0 & 1.0 & 8.0 & 41.0 & 21.0 & \textbf{47.0} & 0.0 & 0.0 & 0.0 & 0.0 & 1.0 & \textbf{10.0} \\
Move Playingcard Away     & 36.0 & 47.0 & 43.0 & 68.0 & 53.0 & \textbf{82.0} & 0.0 & 0.0 & 11.0 & 3.0 & \textbf{22.0} & 21.0 \\
Move Stapler Pad          & 0.0 & 1.0 & 2.0 & 12.0 & 0.0 & \textbf{39.0} & 0.0 & 0.0 & 0.0 & 0.0 & \textbf{2.0} & 0.0 \\
Open Laptop               & 56.0 & 49.0 & 59.0 & 82.0 & \textbf{85.0} & 84.0 & 0.0 & 0.0 & 32.0 & 7.0 & 46.0 & \textbf{69.0} \\
Open Microwave            & 86.0 & 5.0 & 37.0 & 61.0 & 80.0 & \textbf{96.0} & 0.0 & 0.0 & 20.0 & 22.0 & \textbf{50.0} & \textbf{50.0} \\
Pick Diverse Bottles      & 7.0 & 6.0 & 2.0 & 52.0 & 27.0 & \textbf{57.0} & 0.0 & 0.0 & 0.0 & 1.0 & \textbf{6.0} & 5.0 \\
Pick Dual Bottles         & 31.0 & 24.0 & 42.0 & 60.0 & 57.0 & \textbf{91.0} & 0.0 & 0.0 & 13.0 & 1.0 & 12.0 & \textbf{18.0} \\
Place A2B Left            & 1.0 & 2.0 & 3.0 & \textbf{46.0} & 31.0 & 31.0 & 0.0 & 0.0 & 1.0 & 2.0 & 1.0 & \textbf{5.0} \\
Place A2B Right           & 0.0 & 13.0 & 1.0 & \textbf{49.0} & 27.0 & 21.0 & 0.0 & 0.0 & 1.0 & 0.0 & \textbf{6.0} & 4.0 \\
Place Bread Basket        & 6.0 & 14.0 & 10.0 & 26.0 & 17.0 & \textbf{59.0} & 0.0 & 0.0 & 2.0 & 1.0 & 4.0 & \textbf{14.0} \\
Place Bread Skillet       & 7.0 & 11.0 & 5.0 & 19.0 & 23.0 & \textbf{60.0} & 0.0 & 0.0 & 1.0 & 0.0 & 1.0 & \textbf{8.0} \\
Place Burger Fries        & 49.0 & 72.0 & 50.0 & 72.0 & 80.0 & \textbf{96.0} & 0.0 & 0.0 & 27.0 & 18.0 & 4.0 & \textbf{40.0} \\
Place Can Basket          & 1.0 & 18.0 & 19.0 & \textbf{67.0} & 41.0 & 40.0 & 0.0 & 0.0 & 6.0 & 2.0 & 5.0 & \textbf{26.0} \\
Place Cans Plasticbox     & 16.0 & 40.0 & 6.0 & 48.0 & 34.0 & \textbf{87.0} & 0.0 & 0.0 & 5.0 & 3.0 & 2.0 & \textbf{25.0} \\
Place Container Plate     & 72.0 & 41.0 & 78.0 & 86.0 & 88.0 & \textbf{92.0} & 1.0 & 0.0 & 17.0 & 1.0 & \textbf{45.0} & 39.0 \\
Place Dual Shoes          & 9.0 & 8.0 & 4.0 & 13.0 & 15.0 & \textbf{44.0} & 0.0 & 0.0 & \textbf{4.0} & 0.0 & 0.0 & \textbf{4.0} \\
Place Empty Cup           & 61.0 & 37.0 & 56.0 & 65.0 & 37.0 & \textbf{79.0} & 0.0 & 0.0 & 7.0 & 1.0 & \textbf{11.0} & 6.0 \\
Place Fan                 & 1.0 & 3.0 & 12.0 & 36.0 & 20.0 & \textbf{42.0} & 0.0 & 0.0 & 2.0 & 1.0 & \textbf{10.0} & 8.0 \\
Place Mouse Pad           & 0.0 & 0.0 & 1.0 & 4.0 & 7.0 & \textbf{49.0} & 0.0 & 0.0 & 0.0 & \textbf{1.0} & \textbf{1.0} & \textbf{1.0} \\
Place Object Basket       & 15.0 & 15.0 & 33.0 & \textbf{65.0} & 16.0 & 55.0 & 0.0 & 0.0 & \textbf{17.0} & 0.0 & 2.0 & 4.0 \\
Place Object Scale        & 0.0 & 1.0 & 1.0 & 15.0 & 10.0 & \textbf{30.0} & 0.0 & 0.0 & 0.0 & 0.0 & 0.0 & \textbf{1.0} \\
Place Object Stand        & 1.0 & 22.0 & 15.0 & 60.0 & 36.0 & \textbf{73.0} & 0.0 & 0.0 & 5.0 & 0.0 & 11.0 & \textbf{20.0} \\
Place Phone Stand         & 2.0 & 13.0 & 15.0 & 44.0 & 35.0 & \textbf{68.0} & 0.0 & 0.0 & 6.0 & 2.0 & 7.0 & \textbf{15.0} \\
Place Shoe                & 5.0 & 23.0 & 35.0 & 58.0 & 28.0 & \textbf{64.0} & 0.0 & 0.0 & 7.0 & 2.0 & 6.0 & \textbf{10.0} \\
Press Stapler             & 31.0 & 6.0 & 41.0 & 69.0 & 62.0 & \textbf{71.0} & 6.0 & 0.0 & 24.0 & 3.0 & 29.0 & \textbf{31.0} \\
Put Bottles Dustbin       & 27.0 & 22.0 & 21.0 & \textbf{60.0} & 54.0 & 50.0 & 1.0 & 0.0 & 4.0 & \textbf{21.0} & 13.0 & 10.0 \\
Put Object Cabinet        & 15.0 & 42.0 & 33.0 & \textbf{72.0} & 68.0 & 50.0 & 0.0 & 0.0 & \textbf{18.0} & 1.0 & \textbf{18.0} & 10.0 \\
Rotate QRcode             & 1.0 & 13.0 & 50.0 & \textbf{74.0} & 68.0 & 43.0 & 0.0 & 0.0 & 5.0 & 1.0 & \textbf{15.0} & 7.0 \\
Scan Object               & 2.0 & 9.0 & 4.0 & 31.0 & 18.0 & \textbf{48.0} & 0.0 & 0.0 & 1.0 & 1.0 & 1.0 & \textbf{2.0} \\
Shake Bottle              & 74.0 & 65.0 & 74.0 & \textbf{98.0} & 97.0 & 93.0 & 10.0 & 8.0 & 45.0 & 19.0 & \textbf{60.0} & 59.0 \\
Shake Bottle Horizontally & 63.0 & 59.0 & 84.0 & \textbf{100.0} & 99.0 & 92.0 & 4.0 & 18.0 & 51.0 & 25.0 & 51.0 & \textbf{55.0} \\
Stack Blocks Three        & 0.0 & 0.0 & 2.0 & 1.0 & 17.0 & \textbf{44.0} & 0.0 & 0.0 & 0.0 & 0.0 & 0.0 & \textbf{8.0} \\
Stack Blocks Two          & 25.0 & 7.0 & 21.0 & 24.0 & 42.0 & \textbf{55.0} & 0.0 & 0.0 & 2.0 & 0.0 & 1.0 & \textbf{13.0} \\
Stack Bowls Three         & 48.0 & 63.0 & 51.0 & 57.0 & 66.0 & \textbf{67.0} & 0.0 & 0.0 & 17.0 & 5.0 & 24.0 & \textbf{25.0} \\
Stack Bowls Two           & 82.0 & 61.0 & 76.0 & 83.0 & \textbf{91.0} & 89.0 & 0.0 & 0.0 & 30.0 & 6.0 & \textbf{41.0} & 25.0 \\
Stamp Seal                & 2.0 & 2.0 & 1.0 & 18.0 & 3.0 & \textbf{44.0} & 0.0 & 0.0 & 0.0 & 0.0 & \textbf{4.0} & 2.0 \\
Turn Switch               & 5.0 & 36.0 & 35.0 & 46.0 & 27.0 & \textbf{56.0} & 2.0 & 1.0 & 15.0 & 8.0 & \textbf{23.0} & 10.0 \\
\midrule
\textbf{Average}         & 29.7 & 28.0 & 34.5 & 55.2 & 46.4 & \textbf{65.7} & 1.7 & 0.6 & 13.7 & 5.0 & 16.3 & \textbf{22.4} \\
\end{longtable}
}

\end{document}